\documentclass[conference]{IEEEtran}
\IEEEoverridecommandlockouts
\usepackage{cite}
\usepackage{amsmath,amssymb,amsfonts}
\usepackage{algorithmic}
\usepackage{graphicx}
\usepackage{subfigure}
\usepackage{textcomp}
\usepackage{xcolor}
\usepackage{multirow}

\usepackage{algorithm}
\usepackage{algorithmic}
\usepackage{amssymb}

\usepackage{pgfplots}
\usepackage{pdfpages}
\usetikzlibrary{spy}
\usetikzlibrary{backgrounds}
\usetikzlibrary{decorations}
\date{}

\usepackage{tikz}
\usepackage{pgfplots}
\usetikzlibrary{external}

\usepackage{lipsum}

\newcommand\copyrighttext{%
  \footnotesize © 2021 IEEE. Personal use of this material is permitted. Permission from IEEE must be obtained for all other uses, in any current or future media, including reprinting/republishing this material for advertising or promotional purposes, creating new collective works, for resale or redistribution to servers or lists, or reuse of any copyrighted component of this work in other works.}
\newcommand\copyrightnotice{%
\begin{tikzpicture}[remember picture,overlay]
\node[anchor=south,yshift=10pt] at (current page.south) {\fbox{\parbox{\dimexpr\textwidth-\fboxsep-\fboxrule\relax}{\copyrighttext}}};
\end{tikzpicture}%
}

\def\BibTeX{{\rm B\kern-.05em{\sc i\kern-.025em b}\kern-.08em
    T\kern-.1667em\lower.7ex\hbox{E}\kern-.125emX}}
\begin{document}

\title{Computer Vision-Based Guidance Assistance Concept for Plowing Using RGB-D Camera\\
}

\author{\IEEEauthorblockN{Erkin Türköz, Ertug Olcay and Timo Oksanen}
\IEEEauthorblockA{\textit{Technical University of Munich},
\textit{School of Life Sciences}, \textit{Chair of Agrimechatronics}, Freising, Germany \\
\{erkin.turkoz, ertug.olcay, timo.oksanen\}@tum.de}
}

\maketitle

\copyrightnotice

\begin{abstract}
This paper proposes a concept of computer vision-based guidance assistance for agricultural vehicles to increase the accuracy in plowing and reduce driver's cognitive burden in long-lasting tillage operations. Plowing is a common agricultural practice to prepare the soil for planting in many countries and it can take place both in the spring and the fall. Since plowing operation requires high traction forces, it causes increased energy consumption. 
Moreover, longer operation time due to unnecessary maneuvers leads to higher fuel consumption. To provide necessary information for the driver and the control unit of the tractor, a first concept of furrow detection system based on an RGB-D camera was developed. 
\end{abstract}


\section{Introduction}
\label{Introduction}

Plowing is a significant activity in agriculture to increase soil fertility and agronomic productivity \cite{lal2007evolution}. In order to provide optimal circumstances to seeds of plants, the soil should be prepared before planting by turning over the upper layer of it. To this end, the  dense and compact soil is loosen to help drainage and root growth.

With the advancements in agricultural machinery, the task has been conducted by tractor-implement systems. Mechanized plowing of huge fields brings about several drawbacks in terms of environment and operation costs. Plowing is one of the most energy consuming practices in agriculture \cite{moitzi2013energy}. For plowing, a remarkable amount of traction force is needed. This results in high fuel consumption and CO\textsubscript{2} emissions. Unnecessary maneuvers of the agricultural machinery also contribute to high fuel consumption and CO\textsubscript{2} emissions. According to recent investigations \cite{vdma}, tractor driver's attention can be improved significantly by providing assistance. Based on this, there is also a potential of extra diesel savings if unskilled or inexperienced tractor drivers are supported by automated guidance systems. 

Automated guidance in agriculture can greatly improve the quality of cultivation and tillage operation. One important module in automated guidance systems is perception. Understanding the scenes in automated agriculture can be a challenging task due to the unstructured and highly variable environment. For perception and guidance assistance systems, sensors such as, Light Detection and Ranging (LiDAR), laser radar (ladar) and different types of cameras can be used. However, having sensors such as LiDAR may yield high expenses. 

In contrast to highway and street scenarios, there are no lane lines or markings in the agricultural fields, especially in plowing. Moreover, changing lighting conditions and variable soil structure make application of computer vision techniques based on RGB images complicated. In addition, lack of color contrast in most plowing tasks is another issue with color-based furrow detection. 


There are two common ways of plowing with a tractor: in-furrow and on-land plowing. In this study, we had an in-furrow plow. In plowing, paying attention to steering is crucial for inexperienced drivers. Especially, driving the tractor straight can be challenging without departing from the furrow line that is taken as reference.

The key contributions of the paper are listed below.
\begin{itemize}
	\item In this paper, we developed an automated labeling scheme for images of furrows arising through the plowing operation. We used Intel RealSense D435 (depth camera) to capture data. Hence, the captured data contain both 3-channel RGB and 1-channel depth information. The labeling scheme is based on a set of classical computer vision techniques and the depth information of images.   
	\item To generalize the approach, we trained a deep learning model on a dataset of various furrow images annotated by our labeling scheme. 
	\item The approach provides the prototype of a Furrow Departure Warning System, which will increase the accuracy in tillage operations. 
	\item With the proposed guidance assistance framework, the tractor driver is able to localize the tractor better with respect to the furrow by considering the visual feedback.
\end{itemize}

The paper is organized as follows: In Section \ref{Related Work}, the related work is reviewed. Section \ref{Approach} describes the materials and the proposed approach for furrow line detection for plowing in detail. Section \ref{Results} presents the results and remarks on the approach. Finally, Section \ref{Conclusion} concludes the paper and provides future research directions.

\section{Related Work}
\label{Related Work}

Machine vision systems have been increasingly used onboard agricultural machines for both autonomous systems and driver assistance in non-autonomous vehicles \cite{pajares2016machine}. Over the past years, different methods and solutions have been investigated for automatic guidance of agricultural machines using computer vision systems in combination with other technologies. The most common guidance application of machine vision systems in agriculture is detecting the crop rows and alleyways in orchards. 

A method for crop row detection in maize fields using RGB images was presented in \cite{montalvo2012automatic}. The approach is mainly based on image segmentation and double thresholding based on the Otsu’s method \cite{otsu1979threshold}, which allows to separate weeds and crops. Since the agricultural operations are mostly outdoor, lighting conditions based on the weather influence the RGB camera-based tasks in a negative way. In study \cite{romeo2013camera}, accuracy in weed and crop line detection based on images under consideration of uncontrolled illumination conditions is investigated.

An automatic expert system for crop row detection in maize fields based on the color information in images acquired by a vision system was presented in \cite{guerrero2013automatic}. A further version of the guidance system, which was implemented and demonstrated for tractor guidance and control, was presented in \cite{guerrero2017crop}. The vision module in this approach is based on segmentation of crop row images to obtain binary images to distinguish the plants from the soil. In addition to studies \cite{gee2008crop,meng2015development} on crop row detection, \cite{ospina2019simultaneous} dealt with crop mapping integrated to automatic guidance systems of agricultural machinery. This was a modified image analysis method based on \cite{sogaard2003determination} by using RGB images.

Another vision-based application in agriculture is mainly for guidance in alleys. In \cite{subramanian2006development}, a machine-vision method for navigating an autonomous tractor in a citrus grove was developed. For the vision-based guidance, color contrast in the alleyways was used for segmenting the path and the trees. However, such a color contrast is not always clearly available in tillage operations for segmenting the furrow and the tilled soil. In addition to crop row and alleyway detection, navigating an agricultural vehicle by using a furrow line as a reference trajectory has been investigated in \cite{morio2017vision}. However, our work differs from earlier works by using depth maps to detect the furrow line in soil in the absence of color variance. 

Besides approaches using camera, LiDAR-based methods, which works with time-of-flight (TOF)
principle, have been also considered for perception tasks in agricultural fields. These are less sensitive to outdoor lighting conditions. However, they are relatively expensive for mass production. The common task using LiDAR in agriculture is to extract crop lines from a point cloud to navigate an agricultural robot autonomously through plant and crop rows \cite{weiss2011plant,hiremath2014laser,malavazi2018lidar}.

\section{Materials and Methods}
\label{Approach}

\subsection{Setup}

\begin{figure}
		\centering
		\subfigure{{\includegraphics[height=1in,width=0.62\linewidth]{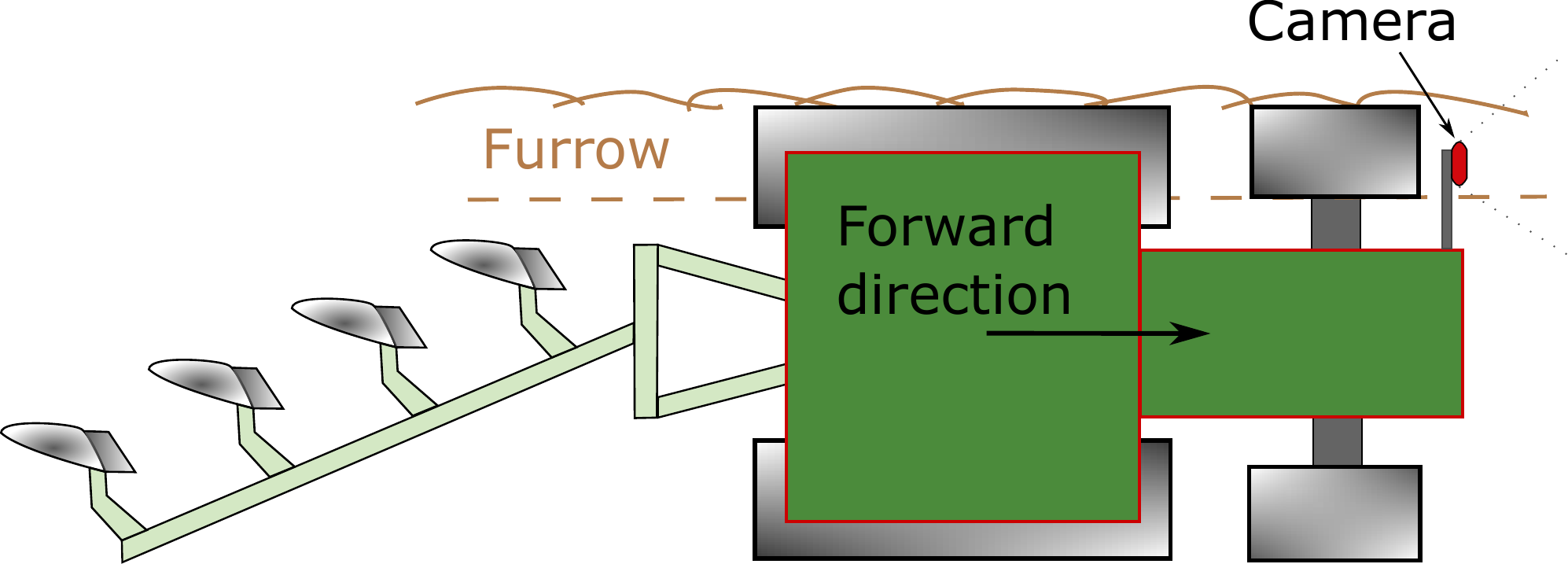} }}%
		\subfigure{{\includegraphics[width=0.38\linewidth]{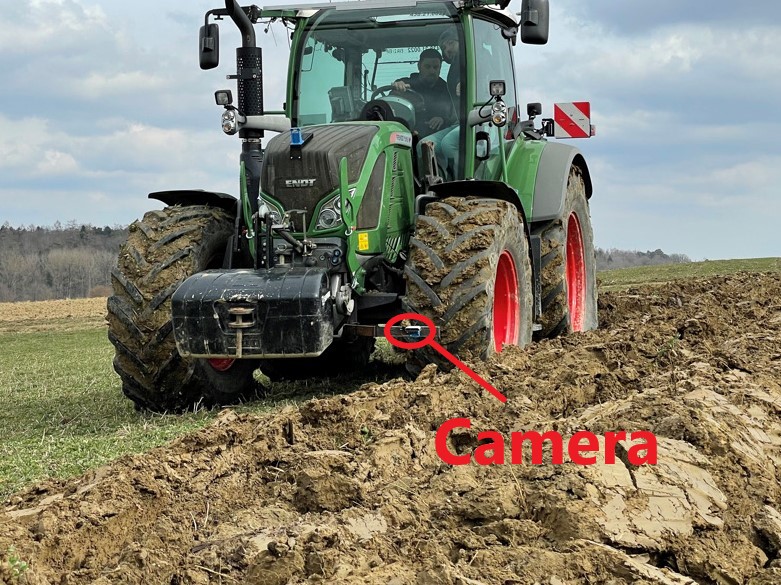} }}%
		\caption{Setup description.}%
		\label{fig:Setup}%
\end{figure}

For this study, we mounted an Intel RealSense depth camera D435 in front of the front wheel, which goes in the furrow. We have taken several captures from three different fields at two different times, November 2020 \& March 2021. We captured images of the furrow, which is usually used as a reference line for the direction of tractor (Fig. \ref{fig:Setup}). The camera focuses on the soil trace from an inclined angle to acquire RGB images with the corresponding colorized depth maps.

The tractor was operated on real terrains showing irregularities and roughness. This yields slight vibrations and swinging in the pitch and roll angles of the vehicle. The camera had pitch and roll angles of -23$^{\circ}$ and 0$^{\circ}$, respectively, at a height of $560$ mm from the ground. The frame rate of the RGB and depth module of the camera was $30$ FPS with resolution of $480 \times 640$ pixels.
By using the data collected with this setup we describe two novel computer vision approaches to detect the furrow edge for guidance assistance purposes.

\subsection{Furrow Detection based on Template Matching} \label{tm}

\begin{figure*}
\centering
\includegraphics[width=0.8\textwidth]{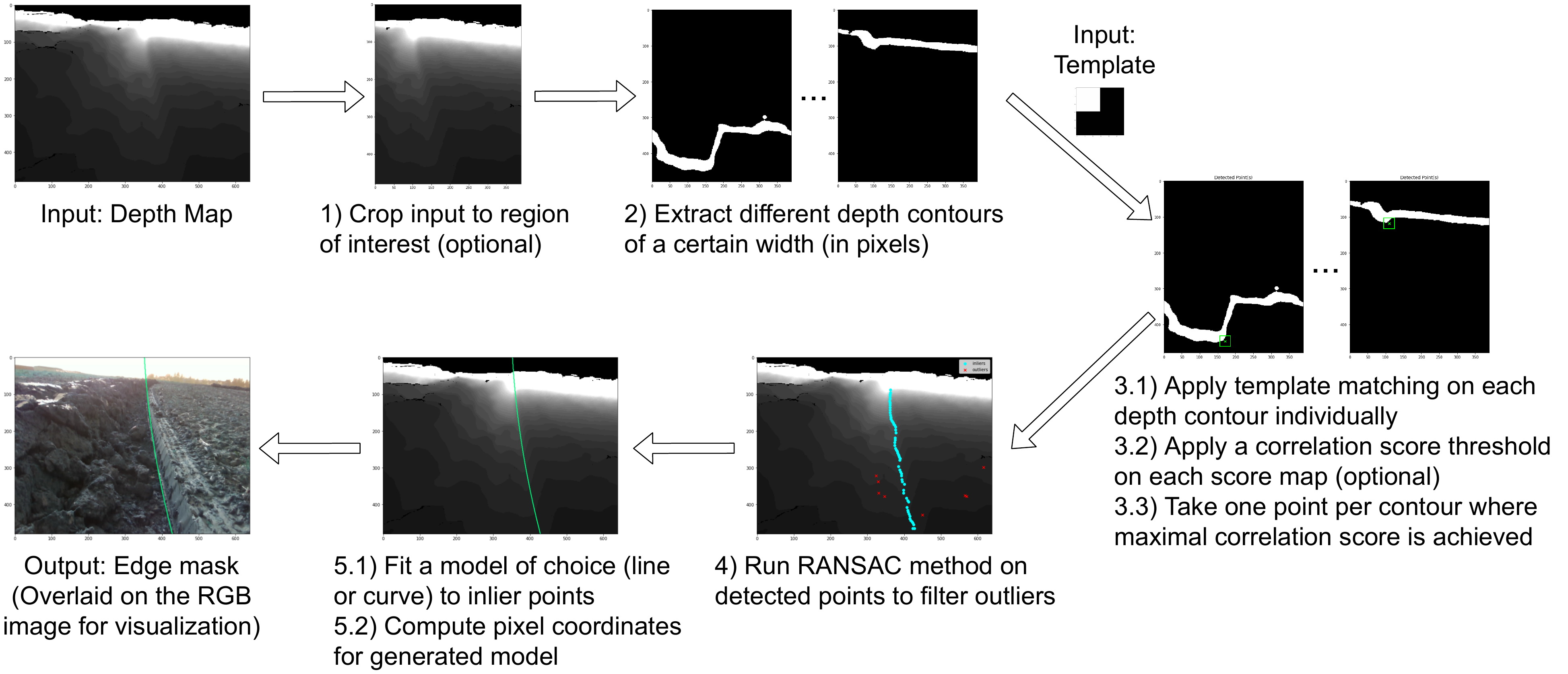}
\caption{Procedure of the proposed furrow edge detection approach based on template matching.}
\label{fig:Procedure}
\end{figure*}

The appearance of the field is affected by the daily weather conditions and the seasonal effects on the vegetation of the field. Such changes pose significant challenges in detecting the edge of the furrow due to a lack of color or illumination variation. Particularly, appearance-based classical approaches for detecting edges on RGB images do not work well in these cases because most of the effective classical methods for edge detection rely on image gradients that indicate abrupt changes in the intensity and color values. For this reason, we turned our attention to the depth map. This depth map is delivered by our RGB-D camera in addition to the RGB image and it is more durable against varying lighting conditions. Viewing this depth map as an image, we can utilize some of the existing computer vision techniques to detect useful features. However, as the gradient-based techniques are not useful on the depth map for our purposes, we propose a method based on template matching. A step-by-step explanation of the proposed method is depicted in Fig. \ref{fig:Procedure}.

The main advantage of this approach is that it is based on the geometry of the furrow but not the appearance of the field. Therefore, it tends to be more robust against varying lighting conditions. However, the major drawbacks of this method are the sensitivity towards corruptions and noise in the depth map and the inability to correctly estimate depth values for the points far away from the camera. Deviation of the edge from the furrow is observable in the distant locations of the sample output depicted in Fig. \ref{fig:Procedure}. Although the robustness of the depth map depends on the quality of the camera, issues with distant points can be avoided to a large extent if we constrain the pitch angle of the depth camera such that it only sees a few meters ahead of the tractor. It is already sufficient to view the part of the furrow edge near the camera to provide an assistance to the tractor.

\subsection{Furrow Detection based on Deep Learning} \label{hed}

Detecting the edge of furrow for a certain environment with the proposed Template Matching method requires tuning a couple of parameters based on the characteristics of that environment in advance. It is an important practical requirement to have a system that does not need any further manual parameter tuning once the development of the system is completed and put into real-life use. Data-driven neural network models are more promising in this regard. Training a neural network model in a supervised manner allows extracting the features from the data automatically and making a prediction based on the extracted features simultaneously. Therefore, when a neural network is trained with a sufficiently large dataset covering most of the possible variations in the environment, it is expected that the system maintains its performance despite environmental changes without further manual adjustments. For this reason, we approached to the same problem with neural networks as well. As our task is fundamentally an edge detection problem, we have chosen a notable neural edge detector, Holistically-Nested Edge Detection (HED) \cite{xie2015holistically}, to present a proof of concept.

\begin{figure*}
\centering
\includegraphics[width=0.9\textwidth]{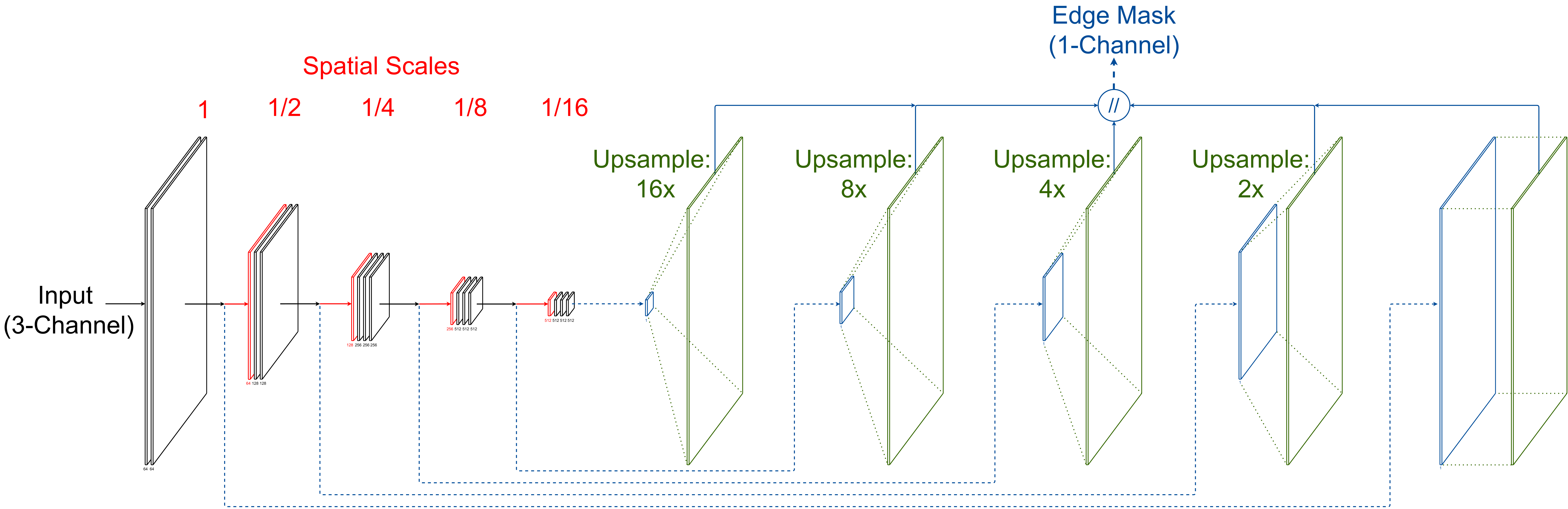}
\caption{HED Architecture: Each rectangle in the figure is a feature map resulting from a layer. Colors denote different layers: black - convolution layer ($3 \times 3$, same) followed by ReLU activation, red - max pooling layer ($2 \times 2$, stride: $2$), blue - $1 \times 1$ convolution layer, green - upsampling layer (no learnable parameters, performs bilinear interpolation). // symbol indicates concatenation operation. There are five side outputs generated from five different scales. The number of channels of each side output is reduced to 1 with $1 \times 1$ convolution. Spatial dimensions of them are upsampled to the input scale. Side outputs are fused with concatenation followed by $1 \times 1$ convolution. Then, sigmoid activation is applied on the fusion map to produce an edge probability mask as the final output.}
\label{fig:hed} 
\end{figure*}

HED is an end-to-end trainable neural network which is designed to perform edge detection. It is built on top of VGG16 architecture \cite{simonyan15vgg} pre-trained on ImageNet dataset, and modified to detect edges on a specific dataset. The network architecture is summarized in Fig. \ref{fig:hed}. The reader is referred to the original paper \cite{xie2015holistically} for further details.

Binary cross-entropy loss is used for training the HED network. However, because of the huge imbalance in the number of edge and non-edge pixels, the loss is balanced between the classes as described in \cite{xie2015holistically}.

\section{Results} 
\label{Results}

We performed experiments on the video recordings taken at two different times, in November 2020 and March 2021. 
While we used the November 2020 captures to develop our proposed methods, we benefited from the March 2021 recordings as the validation and test sets to evaluate the performance. As it is costly, difficult and time-consuming to generate ground truth edge masks for a large amount of data by manual labeling, we used our Template Matching method to extract the edge masks. Our Template Matching method provides overall good but still imperfect annotations. Since they are not good enough to make a quantitative evaluation, we only qualitatively compare the results for the proposed methods and a selected classical method that delivered the best visual results among the methods tested on our data.
\begin{figure}
\centering
\includegraphics[width=0.40\textwidth]{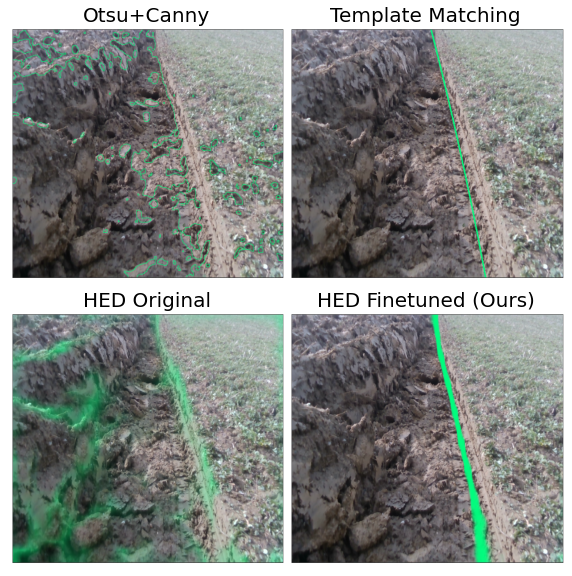}
\caption{Detection visualization with different methods on a sample in the validation set. While Otsu+Canny \& HED Original take RGB image as input, Template Matching and HED Fine-tuned process 1-channel depth map.}
\label{fig:comparison}
\end{figure}

\begin{figure*}
\centering
\includegraphics[width=\textwidth]{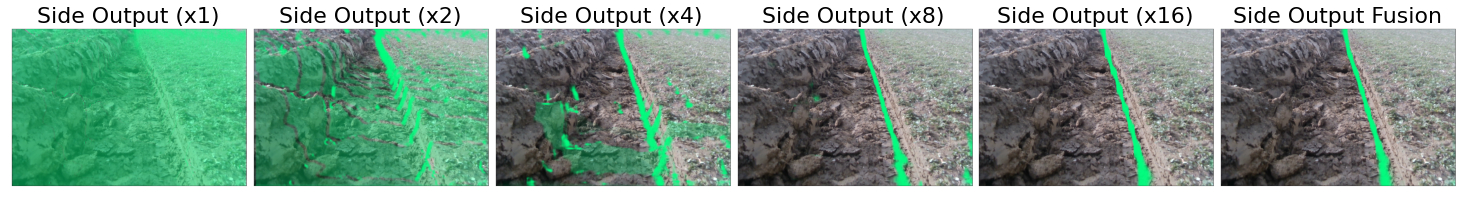}
\caption{Outputs of HED for Depth-Only input. The best detection is in the largest scale (x16). The final output is the fusion of $5$ side outputs.}
\label{fig:variants}
\end{figure*}

We experimented a couple of well-known classical methods with the help of OpenCV and scikit-image libraries. Of the methods and their combinations tested on our data, the best performing model was achieved by using the Otsu Thresholding followed by Canny Edge Detector after some preprocessing. An example result for this classical pipeline on the validation set is shown in the top left of Fig. \ref{fig:comparison}. However, the problem here is that although we can clearly observe the furrow from the output of this approach, we cannot exactly model the furrow edge. Therefore, the result achieved with this method has to be postprocessed so that it can be useful for our purposes. Another drawback is that careful tuning of the parameters is required even for this incomplete result. The top left result shown in Fig. \ref{fig:comparison} is achieved with the following procedure: (1) converting the image to grayscale, (2) applying Gaussian Blur with kernel size $11 \times 11$ and $\sigma=22$, (3) thresholding with Otsu's method, and finally, (4) using the Canny Edge Detector with high threshold equivalent to the threshold returned in step 3, low threshold being the half of the high threshold and a kernel size of $5$ for the Sobel Operator which is used internally in Canny Edge Detector.

\begin{figure}
\centering
\includegraphics[width=0.45\textwidth]{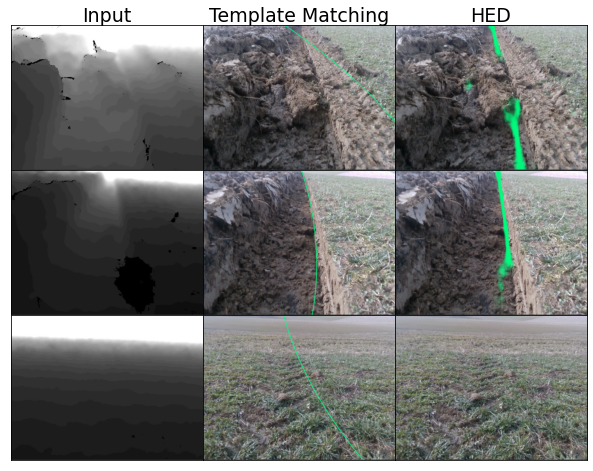}
\caption{Some examples for uncommon cases: (1) soil is piled on the furrow, which occludes the edge, (2) the depth map is corrupted for a reason, (3) the tractor leaves the furrow behind and the edge is not visible anymore. In all cases, Template Matching fails because assumptions related to the depth map are violated. On the contrary, outputs of HED are still reasonable although the performance degrades to some extent.}
\label{fig:tm-hed}
\end{figure}

Since it is difficult to achieve a good result with the existing classical methods for detecting the desired edge on our RGB images, we proposed the Template Matching method for the available depth data. We are able to model a single edge with this approach. However, this method heavily depends on the quality of the depth map. Corrupted depth maps, as the one in the 2nd row in Fig. \ref{fig:tm-hed}, lead to poor results. Moreover, the estimated model cannot fit the distant edge points well because the accuracy of depth values decreases significantly for those points. \footnote{According to Intel RealSense D400 datasheet, depth values are in the range from $0.2$ to over $10$ meters. Camera pose should be adjusted accordingly.} Lastly, unless feasible criteria to reject false detections are manually specified, this method will continuously detect an edge even if there is none. Such criteria are, however, simple heuristics that will most probably not generalize well to different cases.

In all results visualized here with this method, we did not use any criteria to reject possibly false edge detections and we used the following values for the parameters: starting depth: 0.92 (in meters), contour width: 25 (in pixels), contour shift: 5 (in pixels), number of contours: infinity, RANSAC threshold: 30, score threshold: 0, ROI: Crop with width: $[250, 640]$ and full height ($[0,480]$), fit type: $2$nd degree parabola. We used the template depicted in Fig. \ref{fig:Procedure} with a patch size of $30 \times 30$.

The drawbacks of Template Matching method motivated us to attempt to the same problem with the data-driven neural network approach. We have chosen HED network \cite{xie2015holistically} to illustrate the advantages of such an approach. As we require labeled data to train HED, we utilized our Template Matching method to create edge masks acting as pixel-wise label for our images. Since annotations by Template Matching are imperfect, after extracting the edge masks, we performed a visual inspection and removed low quality detections from the training set. Furthermore, we have taken $400 \times 400$ crops from each sample to reduce the possible adverse effect of poor predictions made by Template Matching method in the distant locations in the image on the neural network training.

We also extended our dataset with data augmentations to cover possible variations. Specifically, we augmented our data through random rotations with small angles, horizontal shift and their combinations. In shift augmentation, we allow a small percentage of augmented data not to include an edge which enables network to learn not to detect an edge when there exists none. The positive effect of negative samples in the training data can be seen in the 3rd row of Fig. \ref{fig:tm-hed}.

Results of the original HED \footnote{Used weights for HED Original model is downloaded from:\\ http://content.sniklaus.com/github/pytorch-hed/network-bsds500.pytorch.} were subjectively assessed prior to training the network with our data. It delivers a better visual result around the region containing the furrow edge than the edge detector that we obtained by combining Otsu \& Canny methods (See Fig. \ref{fig:comparison}). However, the default version of HED also provides additional responses in other locations of the image. Redundant responses have to be filtered out with post-processing to make the resulting detection useful for our purposes.

We then trained the same HED architecture as \cite{xie2015holistically} with our own data annotated via our Template Matching method. Training data initially contained frames from $5$ different video captures from November 2020. Furthermore, each training capture is copied and augmented. In this way, $10$ videos are obtained for training in total. For validation and test sets, however, two video recordings per set from March 2021 were used. In the end, train, validation, test splits included 10K, 2K and 2K frames respectively. HED is implemented with PyTorch library and training was performed on Nvidia RTX 2060S graphics card. We used Adam optimizer with learning rate of $5e-04$. Experimentally, we find that using the pre-trained weights of VGG16 and fine-tuning the entire network with our data delivers the best performance.

We trained the network with $3$ input variants: (1) Depth-Only, (2) RGB-Only, (3) RGB+Depth for a maximum of 20 epochs. During training, a batch size of 20 is used for Depth-Only and RGB-Only cases while it is reduced to 16 in RGB+Depth case because of the increasing memory usage with 4 channel input and additional convolutional layer, which is inserted before the first layer to reduce 4 channel input to 3 channels again. While variants including RGB input showed overfitting behaviour in training and delivered poor visual results, Depth-Only input provided satisfactory detections. In Fig. \ref{fig:variants}, side outputs for Depth-Only case in the order of increasing receptive field are depicted. In the right-most column, the final output obtained after fusing $5$ side outputs is shown.

Finally, we compare the performance of two methods proposed in this work. \footnote{We prepared a short video for visual comparison:\\ https://youtu.be/OwIoUGY-RF8.} When the depth map is in a good quality, both methods provide seemingly good results. In such cases, Template Matching is advantageous because it provides single pixel response for each point along the furrow edge. On the other hand, HED provides multiple responses for each point on the edge. Hence, it requires some post-processing to obtain edges with single pixel width. However, in the uncommon cases, such as those in Fig. \ref{fig:tm-hed}, HED performance degrades gracefully while Template Matching fails completely. Although HED is not explicitly trained for such cases, it provides reasonable outputs. These results suggest that HED with Depth-Only input can generalize to uncommon situations contrary to Template Matching approach.
%

Furthermore, while Template Matching is far away from real-time processing, HED is able to produce the edge map prediction from a depth input with a performance close to real-time, on the average around $23$ FPS on graphics card used in this study. Our code is available at: https://github.com/free-bit/furrow-edge-detector

\section{Discussion and Conclusions}
\label{Conclusion}

In this paper, we have proposed two methods to find an edge in the furrow for localization of the tractor. Here, we summarize these methods, propose some ways for improvement, and mention some possible future applications.

It is currently impractical to use the Template Matching method in the real-world applications because it is extremely slow and highly sensitive to parameter settings. Converting it to a fast and reliable method requires non-trivial efforts but the performance will remain to be limited by the quality of the depth map. However, Template Matching shows a good potential to become a useful tool to annotate the data offline with some changes. For example, the points returned by the method can be manually processed to improve the annotation quality particularly in the regions distant to camera.
 
Although HED, the proposed neural network approach, is much faster and seems to be more durable against corruptions in the depth map than the Template Matching method, it still requires some modifications for end use. First of all, the quality of the ground truth annotation provided by Template Matching has to be improved. Then, the performance of the network in edge detection also has to be quantitatively evaluated, e.g. with F1-based Optimal Dataset Scale (ODS) and Optimal Image Scale (OIS) metrics \cite{ois11}. Moreover, a suitable set of post-processing techniques has to be determined and applied to thin the network's final edge prediction.

In our work, we used HED architecture to provide a proof of concept for the applicability of neural networks to this task. However, there is a wide array of architectures that can be used instead. For example, we can replace HED with a U-Net \cite{unet15} entirely, or we can partially replace HED by changing the backbone network from VGG to a ResNet \cite{He2015} variant. Alternatively, existing architecture can be modified by adding different layers such as Batch Normalization or Dropout.

Using a depth map as an input to the neural network is likely to be a disadvantage hindering its wide-scale use because depth cameras are not as commonly used as RGB cameras. If more data with more variation is annotated with better quality, then the network might generalize well to unseen cases also with the RGB input.

\begin{figure}
\centering
\includegraphics[width=0.25\textwidth]{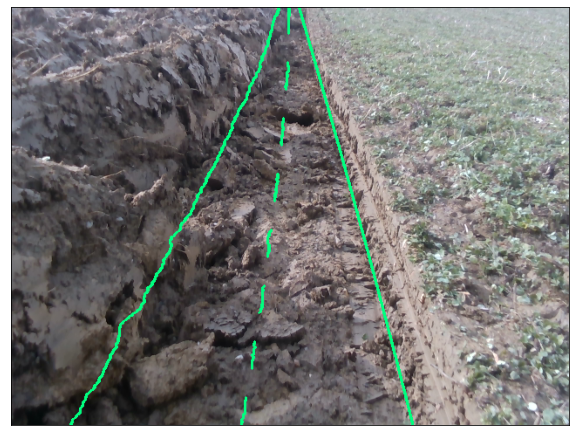}
\caption{Visual Furrow Departure Warning System with Template Matching Method: Artificial lane lines can be drawn to provide a visual feedback to the driver after detecting the edge of the furrow and additionally using wheel width information, which is 530 mm in our case.}
\label{fig:end-result}
\end{figure}

Automatic furrow edge detection introduces useful applications. A target future application is a real-time visual guidance assistance system for furrow departure as illustrated in Fig. \ref{fig:end-result}. A more advanced application is measuring the deviation from the furrow and passing it to the tractor's control system.

As a future work, the temporal information between the frames can be incorporated to make predictions robust against corruptions in the depth map. When the system is real-time capable, it can be deployed and efficiencies in terms of fuel consumption and CO\textsubscript{2} emission between the unguided and guided operations can be compared.

\section*{Acknowledgment}

The authors would like to thank Michael Maier for his technical support in experimental setup for data acquisition, organization of the test fields and plowing.

\bibliographystyle{IEEEtran}
\bibliography{IEEEabrv,Literatur}

\vspace{12pt}

\end{document}